\documentclass{article}
\usepackage{spconf,amsmath,graphicx,hyperref}

\usepackage{amssymb}
\usepackage{amsthm}
\usepackage{multirow}
\usepackage{enumitem}
\usepackage[dvipsnames]{xcolor}

\title{Enhanced Graph Transformer with Serialized Graph Tokens}
\name{
Ruixiang Wang$^{1,2}$ \qquad Yuyang Hong$^{1,2}$ \qquad Shiming Xiang$^{1,*}$ \qquad Chunhong Pan$^{1}$
\thanks{*Corresponding author.}
}
\address{
$^{1}$MAIS, Institute of Automation, Chinese Academy of Sciences \\ 
$^{2}$School of Artificial Intelligence, University of Chinese Academy of Sciences
}

\begin{document}
%
\maketitle
\begin{abstract}
Transformers have demonstrated success in graph learning, particularly for node-level tasks.
However, existing methods encounter an information bottleneck when generating graph-level representations.
The prevalent single token paradigm fails to fully leverage the inherent strength of self-attention in encoding token sequences, and degenerates into a weighted sum of node signals.
To address this issue, we design a novel serialized token paradigm to encapsulate global signals more effectively.
Specifically, a graph serialization method is proposed to aggregate node signals into serialized graph tokens, with positional encoding being automatically involved.
Then, stacked self-attention layers are applied to encode this token sequence and capture its internal dependencies.
Our method can yield more expressive graph representations by modeling complex interactions among multiple graph tokens.
Experimental results show that our method achieves state-of-the-art results on several graph-level benchmarks.
Ablation studies verify the effectiveness of the proposed modules.
\end{abstract}
\section{Introduction}

Graph Neural Networks (GNNs) \cite{bruna2014spectral, gilmer2017neural} have demonstrated remarkable success in modeling graph data from related tasks.
A significant class of tasks, such as predicting molecular properties \cite{sterling2015zinc} or determining protein function \cite{hu2020open}, necessitates reasoning about an entire graph.
These graph-level tasks require GNNs to map an entire graph to a single label or value.
Therefore, it is essential to encapsulate node and edge signals into a comprehensive graph-level representation.

The graph-level representation must be invariant to node permutation and have a fixed dimension, independent of the number of nodes, to serve as valid inputs for the feed-forward network (FFN) on the head.
Beyond these constraints, an effective representation is expected to encapsulate sufficient global signals to be discriminative for downstream tasks.

Nevertheless, existing methods face a challenge in effectively aggregating global signals without creating an information bottleneck.
Naive pooling methods (such as sum or mean \cite{gilmer2017neural}) can satisfy the constraints and be efficient, but suffer from substantial information loss \cite{xu2018powerful}.
Recent works have explored more powerful Transformer architectures to aggregate global signals into node tokens \cite{VaswaniSPUJGKP17}.
Subsequently, to access graph-level representations with fixed dimension, a prevalent paradigm involves introducing a special virtual node connected to all other nodes \cite{ying2021transformers}, or prepending a graph token to the node token sequence \cite{KimNMCLLH22}.
The final embedding of this special node or token, processed through layers of self-attention, is then taken as the graph-level representation.
Under this paradigm, recent methods explore the introduction of spectral signals \cite{BoSWL23}, random walk probabilities \cite{Ma0LRDCTL23}, or subgraph information \cite{Bar-ShalomBM24}, which further enrich the representation.
Although this paradigm allows for more sophisticated, attention-based weighting of node signals than naive pooling \cite{kreuzer2021rethinking, HeH0PLB23}, it still collapses the entire graph signal into a single token.
Consequently, the single token paradigm degenerates into a highly parameterized weighted sum of node signals.

\begin{figure}[t]
  \centering
  \centerline{\includegraphics[width=1.0\linewidth]{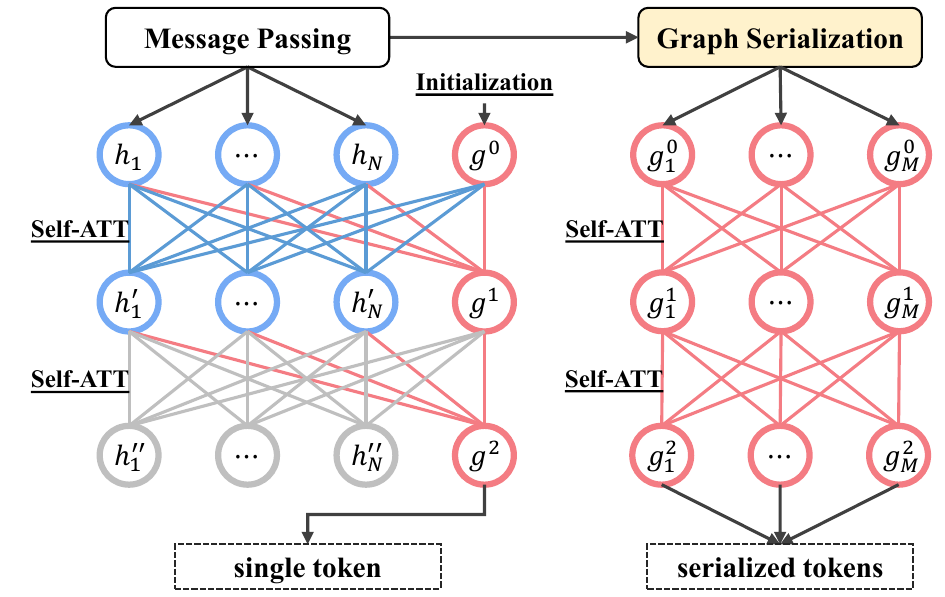}}
\caption{
Paradigms.
The single token paradigm (left) collapses the graph into a single token, which risks over-compression of node signals.
Our serialized token paradigm (right) models the graph as a token sequence to retain more global signals.
}
\label{compare}
\end{figure}

Fundamentally, this paradigm risks over-compression of node signals, which is shown in Figure \ref{compare}.
It underutilizes the core strength of self-attention to model complex interactions within serialized tokens.
With node signals compressed into one token, the rich relational context that could otherwise be captured across the graph is lost.
Some recent works attempt to generate multiple graph tokens, yet naive pooling is still applied as the final output \cite{0003SZL24, buterez2025end}, and further exploration of their intrinsic dependencies remains lacking \cite{HussainZS22, 0003SZL24}.

To this end, we propose a new paradigm to encapsulate global signals more effectively, which reframes graph-level representation learning from single token aggregation to serialization modeling.
As shown in Figure \ref{compare}, node signals are first serialized as a fixed-length graph token sequence.
This sequence is then encoded as the graph-level representation.
Specifically, a graph serialization method is proposed, where a sequence of learnable basis tokens is trained to aggregate node signals into serialized graph tokens and provide positional encodings.
By applying self-attention to encode this token sequence, our method can capture its internal dependencies and encapsulate global signals effectively.
This approach overcomes the information bottleneck of the single token paradigm.
It fully leverages the capacity of self-attention to model complex interactions, thereby yielding more expressive and discriminative graph-level representations.

Our contributions are as follows.

(i) We propose a graph serialization method to generate serialized graph tokens.
It can explicitly preserve more node signals and automatically involve the positional encoding provided by a sequence of learnable basis tokens.

(ii) We design a serialized token paradigm to generate more expressive graph-level representations.
It overcomes the information bottleneck of single token aggregation by modeling complex interactions among multiple graph tokens.

(iii) Experimental results show that our method achieves state-of-the-art results on several graph-level benchmarks.
Ablation studies also verify the effectiveness of our designed serialization module and self-attention module.

\section{Task Formulation}

A graph data is defined as $\mathcal{G}=\left(\mathcal{V}, \mathcal{E}\right)$, where $\mathcal{V}$ is a set of $N$ nodes and $\mathcal{E} \subseteq \mathcal{V} \times \mathcal{V}$ is a set of edges between nodes.
Each node $v_{i} \in \mathcal{V}$ has a label $\boldsymbol{x}_{i}^{v}$ and each edge $e_{ij} \in \mathcal{E}$ has a label $\boldsymbol{x}_{ij}^{e}$.
The one-step neighborhood of node $v_{i}$ is denoted as $\mathcal{N}(i)$.
In graph-level tasks, the target of GNNs is to make a prediction $\boldsymbol{y}_{\text{pred}}$ for the entire graph $\mathcal{G}$.
The prediction can be a discrete label in classification tasks or a continuous value in regression tasks.
The general process begins by generating node-level representations $\boldsymbol{h}_{i}$ for each node, which are then encapsulated into a graph-level representation $\boldsymbol{g}$.
Finally, $\boldsymbol{g}$ is fed into an FFN to output the prediction $\boldsymbol{y}_{\text{pred}}$.

\section{Methodology}

\begin{figure*}[t]
  \centering
  \centerline{\includegraphics[width=1.0\linewidth]{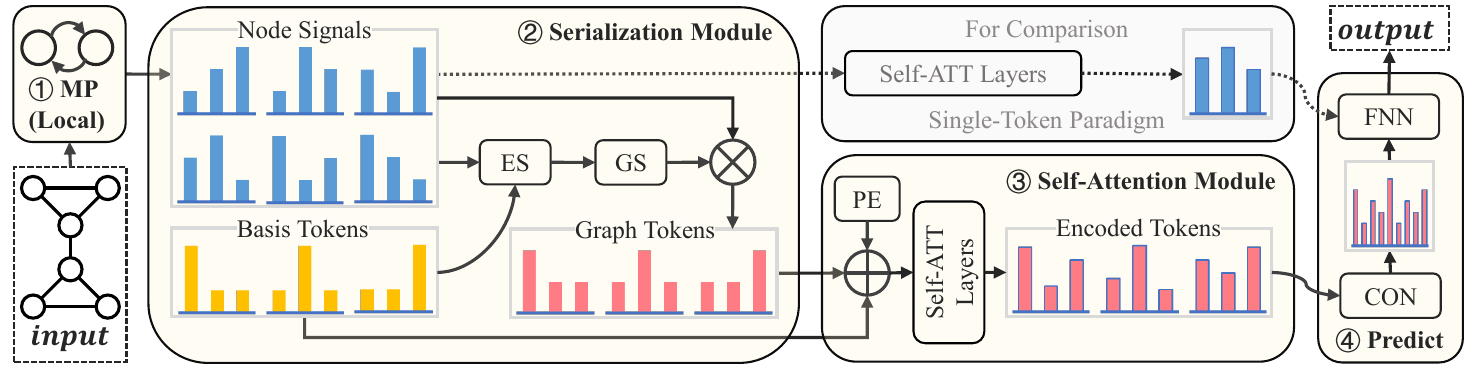}}
\caption{Model structure. Our serialized token paradigm comprises four modules. ``ES'' denotes Euclidean similarity, ``GS'' denotes Gumbel Softmax, and ``CON'' denotes concatenation.
The single token paradigm is also illustrated for comparison.}
\label{model}
\end{figure*}

Under our paradigm, the GNN model comprises four components: a local message passing (MP) module, a serialization module, a self-attention module, and a prediction module.
Figure \ref{model} illustrates the model structure.

\subsection{Embedding and Local Message Passing}

The local MP module aims to generate node-level representations.
This process begins with an embedding layer that encodes the raw labels of nodes and edges into high-dimensional feature vectors, which can be denoted as $\boldsymbol{h}_{i}^{0}$ and $\boldsymbol{e}_{ij}^{0}$:
\begin{equation}
    \boldsymbol{h}_{i}^{0} = \operatorname{embedding}\left( \boldsymbol{x}_{i}^{v} \right), \ 
    \boldsymbol{e}_{ij}^{0} = \operatorname{embedding}\left( \boldsymbol{x}_{ij}^{e} \right).
\end{equation}

Following the embedding layer, a stack of multiple local MP layers is applied.
In each layer, a node updates its feature vector by aggregating local signals.
This process incorporates signals from both the neighboring nodes and the edges that connect them.
The $l$-th layer can be formulated as:
\begin{equation}
    \boldsymbol{h}_i^{l} = f^{l} \left( \varepsilon^{l} \cdot \boldsymbol{h}_i^{l-1} + \sum_{j \in \mathcal{N}(i)} \phi^{l} \left(\boldsymbol{h}_{j}^{l-1}, \boldsymbol{e}_{ij}^{l-1} \right) \right),
\end{equation}
where $\varepsilon$ is a learnable scalar, and $f(\cdot)$ is the update function, typically an FFN.
$\phi(\cdot,\cdot)$ is used to fuse the node and edge signals, which consists of a concatenation and an FFN.
The feature vector of the edge can be updated by fusing its two endpoints: $\boldsymbol{e}_{ij}^{l}=\phi \left(\boldsymbol{h}_{i}^{l}, \boldsymbol{h}_{j}^{l} \right)$.
After iterative MP layers, this module outputs the node feature vectors $\left\{ \boldsymbol{h}_{1}, \cdots, \boldsymbol{h}_{N} \right\}$.

\subsection{Graph Serialization and Positional Encodings}

The serialization module aims to generate serialized graph tokens with positional encoding.
Specifically, a sequence of $M$ learnable vectors $\left\{\boldsymbol{b}_{1}, \cdots, \boldsymbol{b}_{M} \right\}$ is trained as the basis tokens.
The node feature vectors $\left\{ \boldsymbol{h}_{1}, \cdots, \boldsymbol{h}_{N} \right\}$ are aggregated into different graph tokens based on their similarity scores with the basis tokens.
The scores are measured by the Euclidean distance and normalized by the Gumbel Softmax:
\begin{align}
    \label{studentt}
    s_{ij}^{\prime} &= \left(1+\left\|\boldsymbol{h}_{i}-\boldsymbol{b}_{j}\right\|^2\right)^{-1}, \\
    \label{gumbelsoftmax}
    s_{ij} &= \frac{\exp \left(\left(s_{ij}^{\prime}+t_{ij}\right) / \tau\right)}{\sum_{m=1}^M \exp \left(\left(s_{im}^{\prime}+t_{im}\right) / \tau\right)},
\end{align}
where $\tau$ is the temperature coefficient, $t_{ij}$ is the Gumbel noise, and $s_{ij}$ is the normalized similarity score between $\boldsymbol{h}_{i}$ and $\boldsymbol{b}_{j}$.
$\tau$ is set to a small value to prevent the signals from being averaged out, which helps maintain discriminativity.
Then the graph tokens can be calculated as follows:
\begin{equation}
    \boldsymbol{g}_j = \sum_{i=1}^{N} s_{ij} \cdot \boldsymbol{h}_{i}.
\end{equation}
This process is illustrated in detail in Figure \ref{model}.
It should be emphasized that the graph token sequence is invariant to node permutations, which is significant in graph learning.

Since the basis token sequence is manually constructed and ordered, the generated graph token sequence is also ordered.
In addition, each basis token corresponds to a specific position in the feature space.
Through training, these tokens are expected to serve as learnable positional encodings (PEs) that can be implicitly integrated into graph tokens.
Finally, this module outputs serialized graph tokens $\left\{\boldsymbol{g}_{1}, \cdots, \boldsymbol{g}_{M} \right\}$ and corresponding PEs $\left\{\boldsymbol{b}_{1}, \cdots, \boldsymbol{b}_{M} \right\}$.

\subsection{Self-Attention on Multiple Graph Tokens}

The self-attention module is employed to generate the graph-level representation.
Given the graph tokens $\left\{\boldsymbol{g}_{1}, \cdots, \boldsymbol{g}_{M} \right\}$ that encapsulate implicit PEs, we explicitly equip them with knowledge of the token order and position anchor by adding sinusoidal PEs and learnable PEs:
\begin{equation}
    \boldsymbol{g}_{pos}^{0} = (1-\lambda) \cdot \boldsymbol{g}_{pos} + \lambda \cdot \boldsymbol{b}_{pos} + \text{SPE}\left(pos\right),
\end{equation}
where $\lambda$ is a scalar, and $\text{SPE}\left(\cdot\right)$ is a sinusoidal function \cite{VaswaniSPUJGKP17}.

Let matrix $\mathbf{G}^0 = \left[\boldsymbol{g}_{1}^0, \cdots, \boldsymbol{g}_{M}^0 \right]^{\top}$ as the initial input.
In each self-attention layer, the input $\mathbf{G}$ is first projected into queries, keys, and values matrices to calculate scaled dot-product self-attention:
\begin{equation}
    \mathbf{Z} = \sigma\left({\mathbf{G}^{l-1} \mathbf{W}_q^l \left(\mathbf{G}^{l-1} \mathbf{W}_k^l\right)^{\top}} / \sqrt{d_k}\right) \mathbf{G}^{l-1}\mathbf{W}_v^l,
\end{equation}
where $\mathbf{W}_q, \  \mathbf{W}_k, \ \mathbf{W}_v \in \mathbb{R}^{d \times d_k}$ are learnable parameters, and $\sigma\left(\cdot\right)$ is the softmax function.
Then, a token-wise FNN is applied to update the token matrix $\mathbf{G}^{l}$:
\begin{equation}
    \mathbf{G}^{l} = \text{LN}\left(\mathbf{G}^{l-1} + \text{FFN}\left(\text{LN}\left(\mathbf{G}^{l-1} + \mathbf{Z} \right)\right)\right),
\end{equation}
where $\text{FFN}\left(\cdot\right)$ is a standard multi-layer perceptron and $\text{LN}\left(\cdot\right)$ is the layer normalization.
After iterative self-attention layers, this module outputs the graph-level representation $\mathbf{G}$.

The key distinction of our method lies in its structural handling of tokens.
The single token paradigm receives and outputs an unordered, variable-sized set of node-level tokens.
It only retains one special token to provide valid input for the subsequent module, as shown in Figure \ref{model}.
In contrast, our method applies self-attention on an ordered, fixed-length sequence of graph-level tokens.
The final encoded token sequence can be completely fed to the prediction module, which captures more complex interactions within the graph.

\subsection{Prediction on Graph Token Sequence}

This module receives a matrix $\mathbf{G} \in \mathbb{R}^{M \times d}$, which contains an ordered, fixed-length sequence of graph tokens.
$\mathbf{G}$ is first expanded into a vector $\boldsymbol{g}^{\prime} \in \mathbb{R}^{Md}$.
Then, an FFN is applied to output $\boldsymbol{y}_{\text{pred}} = \text{FFN}\left( \boldsymbol{g}^{\prime} \right)$ for the downstream task.

\section{Experiments}

This section reports the experimental results.
We first introduce the benchmarks.
Then the proposed method is compared with recent state-of-the-art methods.
Ablation studies are provided to evaluate the impact of the proposed modules.

\subsection{Benchmarks and Evaluation Procedures}

We evaluate our models on the following three benchmarks.

\textbf{ZINC} is a graph regression benchmark for predicting the constrained solubility of drug molecules \cite{sterling2015zinc}.
It contains 12,000 molecular graphs, divided into training, validation, and test sets according to a widely adopted protocol \cite{DwivediJL0BB23}.
The Mean Absolute Error (MAE) on the test set is evaluated, where lower values indicate better performance.

\textbf{ZINC-FULL} is the full-scale version of ZINC, containing approximately 250,000 molecular graphs \cite{sterling2015zinc}.
It facilitates the development of more expressive models trained on large-scale data, which draws attention from recent research.

\textbf{MolHIV} is a graph classification dataset from the Open Graph Benchmark (OGB) for predicting whether a molecule inhibits HIV replication \cite{hu2020open}.
It contains 41,127 molecular graphs, following the official data splits from OGB.
The Area Under the ROC Curve (AUC) on the test set is evaluated, where higher values indicate better performance.

We train models from scratch on these benchmarks.
During training, the epoch exhibiting the best performance on the validation set is selected.
Each model is trained for four runs using different random seeds.
The mean and standard deviation of the performance on the test set across four runs are reported as the final results.

\subsection{Comparison with State-of-the-Art Methods}

Our method, named Serialized Tokens based Graph Transformer (STGT), is compared with recent advanced methods.
They include both classical paradigms \cite{ying2021transformers, RampasekGDLWB22} and state-of-the-art performance \cite{luo2025can, buterez2025end} in the fields of message-passing neural networks (MPNNs) and graph Transformers.

\begin{table}[t]
\centering
\caption{\label{zinc-hiv}
Performances of recent methods on ZINC and MolHIV.
The top results are marked \textcolor{ForestGreen}{1st}, \textcolor{NavyBlue}{2nd}, and \textcolor{YellowOrange}{3rd}.
${\dagger}$ indicates pre-trained models. N/A denotes results not provided.
}
\vspace{0.05 in}
\begin{tabular}{l|cc}
\hline
Model & \textbf{ZINC} {\footnotesize (MAE $\downarrow$)} & \textbf{MolHIV} {\footnotesize (AUC $\uparrow$)}  \\
\hline
Graphormer \cite{ying2021transformers}  & 0.122 ± 0.006 & 0.8051 ± 0.0053$^{\dagger}$  \\
SAN \cite{kreuzer2021rethinking}        & 0.139 ± 0.006 & 0.7785 ± 0.2470  \\
EGT \cite{HussainZS22}                  & 0.108 ± 0.009 & 0.8060 ± 0.0065$^{\dagger}$  \\
GraphGPS \cite{RampasekGDLWB22}         & 0.070 ± 0.004 & 0.7880 ± 0.0101  \\
Graph ViT \cite{HeH0PLB23}              & 0.073 ± 0.001 & 0.7997 ± 0.0102  \\
GRIT \cite{Ma0LRDCTL23}                 & \color{NavyBlue}{0.059 ± 0.002} & 0.7835 ± 0.0054  \\
Specformer \cite{BoSWL23}               & 0.066 ± 0.003 & 0.7889 ± 0.0124  \\
Cluster-GT \cite{0003SZL24}             & 0.071 ± 0.004 & N/A              \\
Subgraphormer \cite{Bar-ShalomBM24}     & N/A           & \color{YellowOrange}{0.8038 ± 0.0192}  \\
GNNPlus \cite{luo2025can}               & \color{YellowOrange}{0.065 ± 0.004} & \color{NavyBlue}{0.8040 ± 0.0164}  \\
\hline
STGT (ours)                             & \color{ForestGreen}{0.055 ± 0.001} & \color{ForestGreen}{0.8163 ± 0.0038}  \\
\hline
\end{tabular}
\end{table}

\begin{table}[t]
\centering
\caption{\label{zinc-full}
Performances of recent methods on ZINC-FULL.
The top results are marked \textcolor{ForestGreen}{1st}, \textcolor{NavyBlue}{2nd}, and \textcolor{YellowOrange}{3rd}.
}
\vspace{0.05 in}
\begin{tabular}{l|c}
\hline
Model & \textbf{ZINC-FULL} {\footnotesize (MAE $\downarrow$)} \\
\hline
$\delta$-2-LGNN \cite{0001RM20}         & 0.045 ± 0.006 \\
$\delta$-2-GNN \cite{0001RM20}          & 0.042 ± 0.003 \\
CIN \cite{bodnar2021cwn}                & \color{YellowOrange}{0.022 ± 0.002} \\
Graphormer \cite{ying2021transformers}  & 0.036 ± 0.002 \\
GraphGPS \cite{RampasekGDLWB22}         & 0.024 ± 0.007 \\
TokenGT \cite{KimNMCLLH22}              & 0.047 ± 0.010 \\
GRIT \cite{Ma0LRDCTL23}                 & 0.023 ± 0.001 \\
SignNet \cite{LimRZSSMJ23}              & 0.024 ± 0.003 \\
ESA \cite{buterez2025end}               & 0.027 ± 0.001 \\
ESA (PE) \cite{buterez2025end}          & \color{NavyBlue}{0.015 ± 0.001} \\
\hline
STGT (ours)                             & \color{ForestGreen}{0.013 ± 0.0003} \\
\hline
\end{tabular}
\end{table}

Table \ref{zinc-hiv} reports their performance on ZINC and MolHIV.
On ZINC, the STGT reduces the error by 6.8\% compared to the state-of-the-art method \cite{Ma0LRDCTL23}, along with improved stability.
On MolHIV, the STGT not only achieves a margin of 1.23 points over the state-of-the-art methods trained from scratch, but also outperforms two methods that utilize pre-trained models \cite{ying2021transformers, HussainZS22} by 1.03 points.
The results on ZINC-FULL are reported in Table \ref{zinc-full}.
Our STGT outperforms the most advanced model and its enhanced variant \cite{buterez2025end}, demonstrating its capability to improve discriminative power and generalization ability through large-scale data.

It should be emphasized that, in contrast to most graph Transformers, our STGT utilizes the self-attention solely for generating graph-level representations.
Node-level representations are instead produced using a local MP strategy.
This design achieves reduced computational cost while preserving competitive performance.
Experimental comparisons with state-of-the-art graph Transformers and MPNNs show that the proposed serialized token paradigm yields more expressive graph-level representations, even without incorporating global signals during node-level learning.

\subsection{Ablation Study}

\begin{table}[t]
\centering
\caption{\label{ablation}
Ablation results on ZINC and MolHIV.
}
\vspace{0.05 in}
\begin{tabular}{l|cc}
\hline
Model & \textbf{ZINC} {\footnotesize (MAE $\downarrow$)} & \textbf{MolHIV} {\footnotesize (AUC $\uparrow$)}  \\
\hline
full model          & 0.055 ± 0.001         & 0.8163 ± 0.0038  \\
w/o serialization   & 0.074 ± 0.003         & 0.7934 ± 0.0151  \\
w/o self-attention  & 0.075 ± 0.002         & 0.7963 ± 0.0096  \\
w/o both            & 0.082 ± 0.004         & 0.7760 ± 0.0214  \\
\hline
\end{tabular}
\end{table}

We conduct ablation studies on both ZINC and MolHIV to validate the effectiveness of the proposed modules.
Specifically, three model variants are constructed:
(i) The serialization is removed, and the self-attention is applied directly on node representations.
(ii) The self-attention is removed, and an FFN is applied directly on the concatenated graph token sequence.
(iii) Both modules are removed, and all node representations are summed as the graph representation.

The results are reported in Table \ref{ablation}.
The removal of any proposed module results in noticeable performance degradation.
Among them, variant (i) represents the single token paradigm, which performs significantly worse than our full model.
This result emphasizes the advantage of our serialized token paradigm in modeling complex interactions.
Variant (ii) adopts a lightweight model without self-attention, which achieves results comparable to the single token paradigm.
However, the computational reduction compared to our full model is limited, as the sequence length remains fixed and is substantially smaller than the node number in our method.
Variant (iii) corresponds to the traditional readout method, which achieves the lowest performance.

In addition, the gain achieved by using both modules together exceeds the sum of gains obtained from using each module individually.
It verifies that the two modules work synergistically, bringing a greater overall improvement.

\section{CONCLUSION}

We have proposed a serialized token paradigm to generate more expressive graph-level representations.
It overcomes the bottleneck of the single token paradigm by modeling complex interactions among multiple graph tokens.
Our method achieves state-of-the-art results on several benchmarks.
The proposed modules are verified to be effective.

\noindent \textbf{Acknowledgment:} This work was supported by the National Natural Science Foundations of China (Grant No.62306310).

\vfill\pagebreak

\bibliographystyle{ieeetr}
\bibliography{strings,refs}

\end{document}